# An algorithm applied the Turing pattern model to control active swarm robots using only information from neighboring modules.

Takeshi Ishida[1]
1 Department of Ocean Mechanical Engineering, National Fisheries University, Shimonoseki, Yamaguchi, Japan
2-7-1, Nagata-honmachi, Shimonoseki, Yamaguchi, 759-6595, Japan

*Corresponding author
 Takeshi Ishida
 E-mail: ishida@fish-u.ac.jp

ORCID of the author(s);

Abstract
   Swarm robots, inspired by the emergence of animal herds, are robots that assemble a large number of modules and self-organize themselves to form specific morphologies and exhibit specific functions. These modular robots perform relatively simple actions and controls, and create macroscopic morphologies and functions through the interaction of a large number of modular robots. This research focuses on such self-organizing robots or swarm robots. The proposed algorithm is a model that applies the Turing pattern, one of the self-organization models, to make a group of modules accumulate and stay within a certain region. The proposed method utilizes the area within the spots of the Turing pattern as the aggregation region of the modules. Furthermore, it considers the value corresponding to the concentration distribution within the spotted pattern of the Turing pattern model (referred to as the potential value in this research), identifies the center of the region (spotted pattern), and makes it the center of the module group. By controlling the modules in the direction of the higher potential value, it succeeds in maintaining the shape of the module group as a whole while moving. The algorithm was validated using a two-dimensional simulation model. The unit module robot was assumed to have the following properties: 1) limited self-drive, 2) no module identifier, 3) information exchange only with adjacent modules, 4) no coordinate system, and 5) only simple arithmetic and memory functions. Using these modules, the devised algorithm was able to achieve not only the creation of static forms but also the realization of the following movements: 1) modules accumulate and grow, 2) modules move to the light source, 3) exit the gap while maintaining its shape, and 4) self-replication. It was demonstrated that a group of robots can be maintained by a basic process in which the module itself calculates potential values, and the module moves based on these values.
Keywords; Swarm robots, Control algorithm, Turing pattern model, Self-organization, Cellular automata

# 1. background

Inspired by the emergence of animal herds, numerous studies have been conducted previously on robots that assemble a large number of modules and self-organize to form specific morphologies and exhibit specific functions [1-3]. The robots in these studies are modular robots that perform relatively simple actions and control tasks, and generate macroscopic morphologies and functions through the interaction of a large number of modular robots. This research focuses on such self-organizing robots, also known as swarm robots. The range of applications of swarm robots utilizing such a self-organizing approach is extensive. Examples include robots that automatically construct structures to adapt to their environment [4] and swarm robots for mapping, exploration, and environmental monitoring [5]. More efficient drug targeting and delivery could be achieved if self-assembling micro- and nanomodules could be constructed [6, 7].

The research lineage of self-organizing robots composed of numerous modules can be categorized based on the mobility of the modules, the number of modules involved, and other factors.

## 1) Classification Based on Module Mobility[8]

Static self-assembling robots, which lack self-propulsion capabilities, rely on external forces to manipulate individual modules, enabling them to aggregate and form crystalline structures through collisions and other interactions. Representative examples of static self-assembling robots include: **Hosokawa's Magnet System[9]:** This system utilizes triangular modules embedded with magnets and explores the process of module assembly into hexagonal structures when agitated within a two-dimensional plane. Hosokawa's work laid the foundation for quantitative analysis in the field of modular robotics. **Eric Klavins' Mechatronic Self-Assembly System[10]:** This system integrates Hosokawa's magnetic blocks with microprocessors and motors. The motors control the magnetic coupling between blocks and facilitate communication among them. Random air currents from the floor propel the blocks, leading to repeated module connections and disconnections. This dynamic process, governed by specific rules, enables the formation of desired shapes. **Griffith's Self-Replicating System[11]:** This system employs pneumatic forces to randomly move a large number of square modules, aiming to create a system capable of replicating a pre-assembled row of modules. Upon collision, modules are joined via a ratchet mechanism located on each side. This connection triggers inter-module communication. The system is designed to minimize the number of module types and the number of states each module can possess.

Dynamic self-organizing robots represent a fascinating class of autonomous systems where individual modules possess self-propulsion capabilities and can autonomously assemble into complex structures as they move. Here delves into several notable examples of research in this domain. **White's Platinum-Powered Modules:** White's study [12] introduced a novel concept of self-driven modules powered by oxygen bubbles. Platinum fins were attached to semi-circular tiles and floated in hydrogen peroxide water. The resulting movement of the tiles, guided by capillary forces, led to the formation of stable pairs. **Clavins' Graph Grammar-Based Coupling Control:** Clavins' work [13] explored the application of graph grammars for controlling coupling between self-driven modules. This approach enabled precise coordination and assembly of modular structures. **Simulations of Self-Driven Particle Properties:** A significant body of research has focused on investigating the properties of self-driven particles through simulations [14]. These simulations provide valuable insights into the collective behavior and emergent patterns of self-organizing systems.

### 2) Classification Based on the Number of Modules

Early research in swarm robotics focused on systems with a relatively small number of modules. These early swarm robots, despite their limited number of modules, often exhibited sophisticated functionalities, such as microcomputers and communication capabilities integrated within individual modules [8-14].

In recent years, however, swarm robots have evolved to incorporate a significantly larger number of modules, ranging from hundreds to thousands[15]. A notable example is the work of Radhika Nagpal's group [16-18], which successfully demonstrated self-organized morphogenesis, including star formation, using approximately 1000 simple robots known as Kilobot. A subsequent study from the same group extended this approach to create swarm shapes through "disassembly" of the swarm using a light attraction/repulsion system [19]. Furthermore, Slavkov et al. utilized hundreds of Kilobot to emulate virtual diffuse matter between robots and solve the Turing pattern equation, resulting in the emergence of specific shapes without prior design [20]. Research in the field of soft matter and other related areas also explores the control of even more active matter, involving tens of thousands or more individual units, through physicochemical interactions [21, 22].

This research delves into the development of control algorithms for self-driven "dynamic" modules with "many modules (tens to hundreds)" within the context of existing research lineages. The realm of self-organizing modular robotics has witnessed the proposal of numerous such robots, each employing diverse algorithms within individual modules. These algorithms can be classified based on the following perspectives:

**1) Distinction between Modules with Unique IDs or Homogeneous Modules without IDs**

Modules equipped with unique IDs facilitate easier control as they enable the selection of a leader module as the control base.

**2) Communication Range**

This classification distinguishes between modules that can only communicate with adjacent neighbors or those that can communicate over longer distances. The specific range of communication between modules determines the classification. Additionally, the ability to determine module individuality based on communication capabilities plays a role in this classification.

**3) Sharing of Coordinate System and Orientation among Modules**

This aspect considers whether each module is assigned an absolute coordinate system, whether relative coordinates are established from a designated module, or whether a coordinate system is entirely absent. Some studies have demonstrated the ability of modules to self-triangulate and establish their own coordinate systems [23].

**4) Shared Timekeeping among Modules**

This classification differentiates between synchronized and shared or asynchronous timing of movements and communications.

When all modules possess unique IDs, can communicate with distant modules, share a common coordinate system aligned with the desired shape, and operate in a synchronized manner, coupled with sufficient computing power in each module for control and information exchange, it becomes feasible to construct the intended shape and motion without excessive reliance on self-organization phenomena. Currently, within the realm of swarm robotics closely aligned with industrial applications, robots are collaborating in this direction to develop functionalities [24, 25].

In contrast, when all modules are homogeneous without unique IDs, can only communicate with immediate neighbors, and lack a shared coordinate system and clock, achieving specific shapes and coordinated movements becomes a significantly more challenging task. A prime example of such self-organizing robots is Harvard University's Kilobot [16-18]. These modules lack unique identifiers, can communicate with neighboring modules, and operate asynchronously. However, they are provided with a set of initial modules and maintain a coordinate system based on design data for the target shape. Subsequently, a specific shape is generated through the superposition of simple rule actions on each Kilobot module. Similarly, Slavkov et al. [20] employed the same Kilobot case study, but replicated the reaction-diffusion model by diffusing virtual substances through inter-module communication, enabling a group of modules to generate a Turing pattern. This serves as another example of a system that does not maintain IDs, relies on communication with immediate neighbors, operates asynchronously, and lacks a shared coordinate system.

When unique identifiers are not assigned to modules, several control methods can be employed to coordinate their behavior:

**1) Schemes based on specific combinations of operation rules:** These methods utilize predefined rules that govern the interactions between modules, leading to the desired collective behavior [26-28].

**2) Schemes based on the number of surrounding modules:** In these approaches, modules make decisions based on the number of neighboring modules they detect, enabling them to adapt their behavior to their local environment [29].

**3) Schemes based on attraction/repulsion:** These methods employ attraction or repulsion forces between modules, guided by simple signals or gradients of state quantities, to achieve desired formations or movements [30-34].

**4) Edge following:** In this approach, modules follow the edges of other swarming robots, enabling them to trace specific paths or patterns [35-37].

**5) Complex interactions:** More sophisticated control methods involve complex interactions between modules, such as:

- Stepwise pattern formation algorithms based on local information for multiple autonomous robots using reaction-diffusion (RD) systems [38].
- Digital hormone models that mimic the behavior of biological hormones to regulate collective behavior [39].
- Gene regulatory network emulation, where modules interact based on gene expression patterns to achieve coordinated behavior [40-41].
- Swarm chemistry, where virtual chemicals are exchanged between modules to influence their behavior [42].

In previous studies, systems lacking unique IDs, restricted to communication with immediate neighbors, and operating asynchronously were found to be limited to generating static shapes, even when retaining design data and incorporating self-organization principles.

Maintaining the desired shape of moving swarm systems proved challenging due to the inherent constraints.

This study introduces a novel algorithm for swarm module robots, validated through a two-dimensional simulation model. The proposed algorithm empowers individual module robots, possessing limited capabilities, to achieve dynamic behaviors beyond static shape formation.

**Unit Module Robot Characteristics:**

1. **Limited Self-Drive:** Module movement is restricted to connections with neighboring modules.
2. **Absence of Unique Identifiers:** Modules lack unique identifiers or designated leaders.
3. **Localized Information Exchange:** Information sharing is limited to neighboring modules.
4. **Non-existent Coordinate System:** Modules do not maintain a shared coordinate system.
5. **Basic Computational Capabilities:** Modules possess only simple arithmetic and memory functions.

Despite these minimal capabilities, the proposed algorithm enables each module to perform the following dynamic behaviors:

1. **Module Accumulation and Growth:** Modules can accumulate and grow in size.
2. **Light Source Navigation and Encapsulation:** Modules can navigate towards a light source while maintaining their accumulated structure, eventually encircling the light source.
3. **Shape-Preserving Gap Exit:** Modules can exit a gap while preserving their overall shape.
4. **Self-Replication:** Modules can replicate themselves through self-assembly.

This study presents a novel algorithm that leverages the Turing pattern, a self-organization phenomenon, to enable a group of modules to accumulate and maintain their position within a designated region. While the robotic system in [20] generates a Turing pattern within a group of modules, our approach utilizes the area within the Turing pattern's spots as the aggregation region for the modules. Furthermore, I introduce the concept of "potential value," which represents the concentration distribution within the spotted pattern of the Turing pattern model. By identifying the center of this spotted pattern, which serves as the center of the module group, and guiding the modules towards areas of higher potential value, I successfully maintain the overall shape of the module group while it moves.

In lieu of directly solving the Turing model, we employ a "temporal model" inspired by the cellular automaton model proposed by Ishida [43] to facilitate communication between modules. This approach eliminates the need for information from remote locations and effectively generates

Turing-like patterns through the exchange of integer values with neighboring cells and the evolution of these values over time. (For detailed explanations, please refer to the Model section.)

Figure 1 provides a visual representation of the proposed algorithm's key components.

1. **Individual Potential Calculation:** Each module independently determines its potential value. This calculation employs Ishida's longitudinal model [43], enabling potential estimation through interactions with neighboring cells only.
2. **Maintaining Position and Connectivity:** Modules maintain their positions within their potential regions, defined as areas with potential values exceeding zero. Simultaneously, they adjust their movement to preserve connections with neighboring modules.
3. **Forward Movement Determination:** To ascertain whether a module is advancing, a positive force acting towards an adjacent empty space is considered. This force is calculated based on the combination of three virtual forces: the force towards the direction of high potential value, the force towards a light source (if present), and a randomly applied force. (These virtual forces are internal calculations within the module to determine direction, not actual forces acting upon the module itself.)

By implementing these rules for each module, the model successfully achieved group cohesion while individual modules navigated without relying on a complex coordinate system. Simulations demonstrated that individual modules could exchange information with neighboring modules and employ a simple algorithm to achieve the desired behavior. While building a physical modular robot to validate the algorithm's effectiveness is the next step in this research, the proposed algorithm is summarized and presented in this paper.

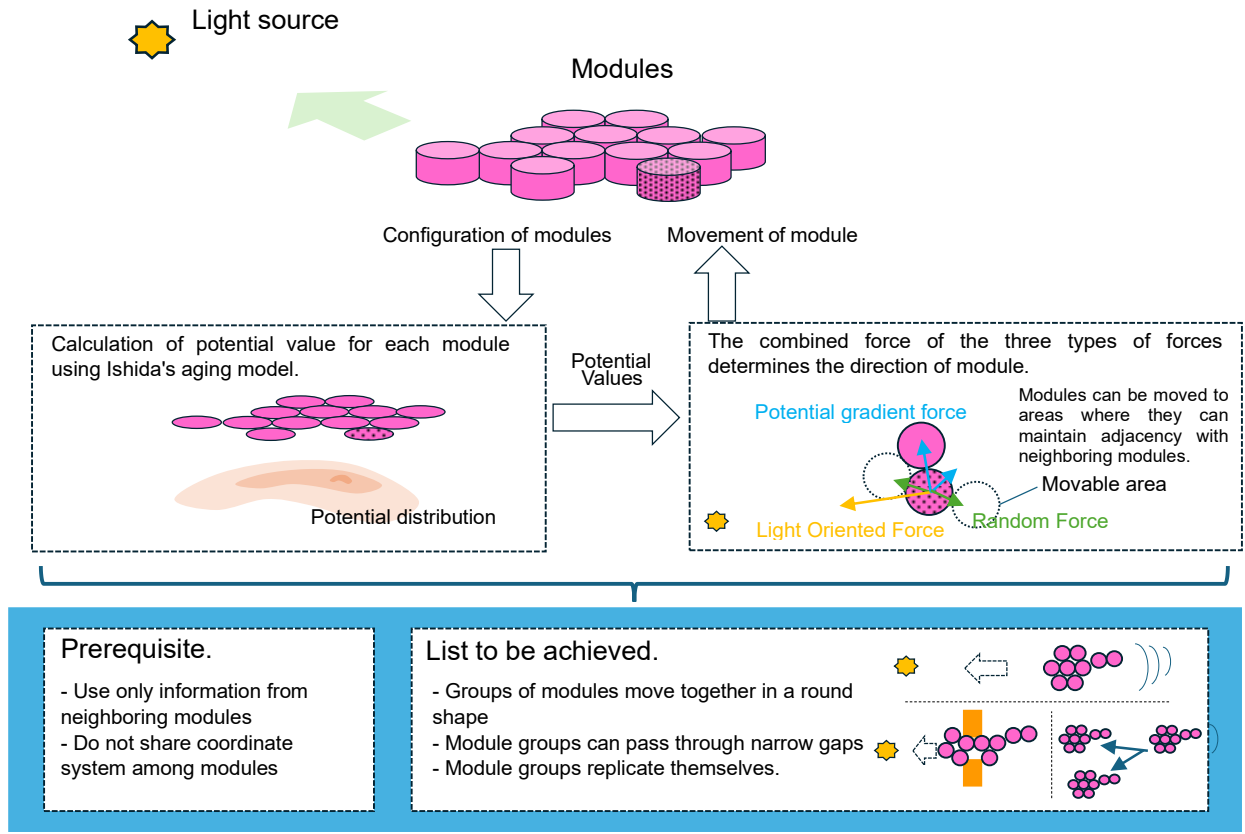

Figure 1: Schematic Diagram of the Proposed Algorithm

## 2. model

The proposed algorithm utilizes Ishida's numerical value aging model [43] to determine the potential region, which in turn guides the movement of modules based on the resultant force exerted by the attractive force arising from the potential value and other relevant forces.

### 2.1 Overview of Ishida's numerical value aging model Model

Ishida's numerical value aging model stands out from conventional Turing pattern generation approaches by employing the exchange of integer values, or "tokens," that evolve over time, rather than relying on two types of diffusive substances. Figure 2 provides a visual representation of the model's core principles.

Multiple tokens are emitted from a black cell and distributed to its neighboring cells. With each movement to an adjacent cell, the token value increases by one. Following a predetermined time interval, the model tallies the number of tokens in each cell. Based on the relative counts of smaller and larger tokens, the model determines the state of the cell in the next iteration. Smaller tokens represent substances with low diffusion coefficients, while larger tokens represent substances with high diffusion coefficients. This unique approach enables Ishida's model to generate intricate patterns by capturing both near and far-reaching interactions.

The detailed workings of Ishida's numerical value aging model model are outlined below (Figure 2A provides a visual overview).

**Assumptions:**

Cells within a two-dimensional hexagonal lattice structure can exist in two states: black and white.

**Token Distribution and Movement:**

1. Black cells emit a specified number of tokens, each initialized with a value of 1.
2. These tokens remain in the originating cell for a proportion determined by the residual rate (two parameters are defined: the calculated amount of tokens and the residual rate).
3. When a token is moved to an adjacent cell, its value is incremented by 1.
4. Token movement is assumed to occur over a maximum of 16 steps.
5. All tokens distributed to neighboring cells are assumed to have equal values. In other words, the total number of tokens, excluding those remaining in the original cell, is divided equally among the six neighboring cells.

**State Determination:**

After token movement has been completed for all cells, the ratio of integer token values in each cell is used to determine whether the cell will be black or white in the next iteration, as per Equation (1):

$$\text{Next cell state} = \begin{cases} \text{Black: (total amount of integer value tokens from 1 to X/2)} > \text{(total amount of integer value tokens from 1 to X)} \times w \\ \text{White: (total amount of integer value tokens from 1 to X/2)} < \text{(total amount of integer value tokens from 1 to X)} \times w \end{cases}$$

-----(1),

**w** represents a weight parameter that influences the transition between black and white states.

Equation (1) governs the transition between black and white states in Ishida's model. The expression essentially determines whether the total number of tokens with values ranging from 1 to **X/2** exceeds the total number of tokens with values ranging from 1 to **X**, weighted by a parameter **w**.

Ishida's model, with its reliance on local interactions, aligns well with the concept of modular robots exchanging information between adjacent modules. The black cells in the model can be envisioned as individual modular robots, and the exchange of token values between adjacent modules can be used to determine whether a particular module is within or outside a Turing pattern. This capability could be harnessed to control the behavior of modular robots in a cooperative manner.

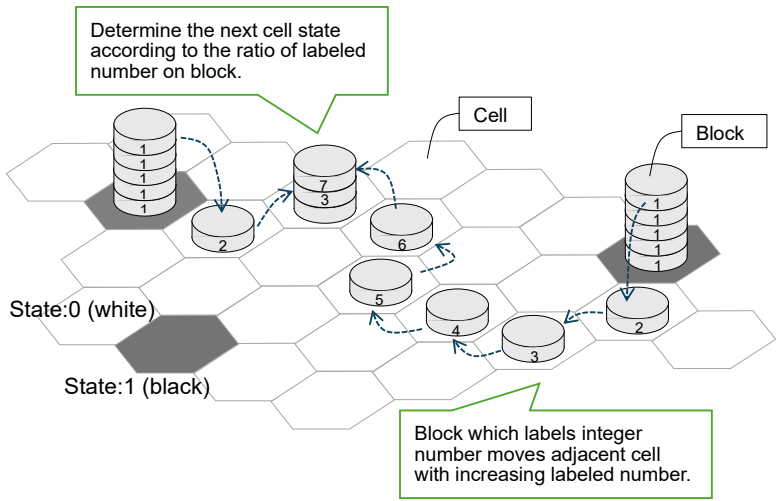
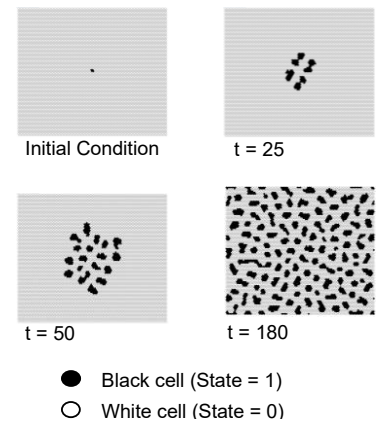

Figure 2: Schematic Representation of Ishida's numerical value aging model Model : This figure depicts a two-dimensional hexagonal grid populated with two-state cells (black and white). **A:** The illustration demonstrates the modification process of tokens generated within the black cells. The token values are incremented up to **X** times within each cell, while some tokens remain unmodified in their respective cells. The frequency distribution of token numbers ultimately determines the subsequent state of each cell. The variable **X** represents the maximum value that can be associated with a token and is a positive integer. Figure Source: [43]. **B**: Example of calculation that generated a Turing pattern.

## 2.2 Simulation Space for Modular Robots

In this study, a two-dimensional hexagonal cellular space was employed, with each cell representing the location of a single modular robot. Each module was programmed to move discretely to an adjacent cell based on the proposed algorithm (Figure 3). The cellular space was configured as a 100 x 100 computational grid.

The decision to utilize a discretized space was twofold: to simplify computational requirements for demonstrating the potential of the proposed robot control algorithm and to maintain consistency with the underlying Ishida's numerical value aging model. Simulation of module movement in continuous space is envisioned as a future research endeavor.

In certain computational scenarios, a designated cell within the space was set as the light source. It was assumed that light is emitted from this source cell in all directions, with its intensity (set to 1 at the source) decaying inversely proportional to the distance. Additionally, in some calculation cases, a continuous wall was introduced within the computational space to create a narrow gap. It was assumed that no module could traverse this wall.

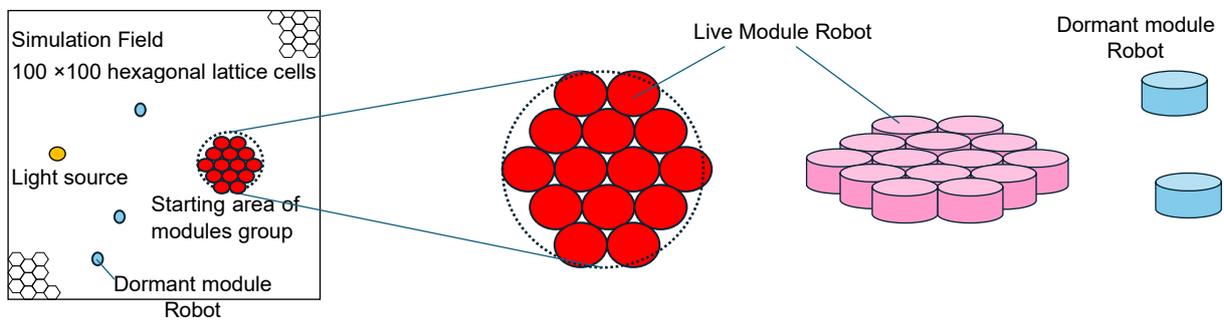

Figure 3: Overview of the Simulation Space for a Modular Robot

## 2.3 Module Control Algorithm

The modules deployed within the simulated environment were endowed with the following functionalities:

1. **Inter-Module Attraction:** A force of attraction exists between modules, drawing them closer and maintaining their adjacency.
2. **Movement:** Modules possess the ability to navigate to adjacent open spaces, excluding walls, while preserving their proximity to neighboring modules.
3. **Light Sensing:** Each module can perceive the direction of the light source.
4. **Information Exchange:** Modules can engage in the exchange of integer values with adjacent modules.

The modular robot was assumed to exist in two distinct states: "active" and "dormant." In the active state, the module robot would engage in light sensing, information exchange with neighboring cells, self-direction determination, and movement. During the dormant state, the robot would remain inactive and perform random movements within the space. The placement of active and dormant modules within the computational space was governed by specific conditions based on the computational scenario, and their movement was guided by the algorithm outlined below. While Ishida's model effectively distributed tokens across the entire space, this model restricts the exchange of numbers (tokens) to a localized region within the module. To address this limitation, the proposed algorithm introduces a slight modification to Ishida's model.

Each module undergoes a repetitive cycle consisting of the following seven steps, as illustrated in the flowchart presented in Figure 4.

1. **Initialization:** All modules begin with an initial integer value of 1.
2. **Information Exchange:** Each module gathers information from its neighboring modules regarding the integers and their quantities held by each neighbor. They then add 1 to the received integers and store the updated count. This exchange is performed simultaneously by all modules. For instance, if a module receives "5 integer 2s" from a neighbor, it stores "5 integer 3s." If other adjacent modules also contribute "1 integer 2," they are added to the existing count of " 2 integer 2s." This process is repeated 20 times for each module in a round-robin fashion. Following this mutual exchange, potential values are calculated using the following formula based on the distribution of integer counts stored by each module (for integers ranging from 1 to 8 and 1 to 16).

    Potential Value = (Total number of integers from 1 to 16) −
    (Total number of integers from 1 to 8) × w  (2)

3. **Potential Area Determination:** The potential area is defined as the region where the potential value is greater than 0. If the potential value is positive, the module remains active. Conversely, if the potential value is negative, the module becomes dormant (during dormancy, the module moves randomly away from the cluster).
4. **Sensing:** Each module gathers information about the directions of adjacent modules and identifies vacant cells that can be accessed while maintaining adjacency with other modules.
5. **Force Calculation:** Modules calculate the attraction between themselves based on their own potential values and those of their neighbors (a greater difference in potential values indicates a stronger attraction). Additionally, the gravitational force exerted by the light source is calculated. Each force is summed by considering its directional component relative to the direction of the empty cell. It is crucial to note that these forces are not actual physical forces acting on the module but rather computational constructs used to determine the likelihood of movement.

**Total Force =** Attraction Force by Adjacent Modules + Attraction Force toward Light Source + Random Attraction Force     (3)

6. **Movement Decision:** If the calculated force in the direction of the vacant cell is positive, the module moves in that direction; otherwise, it remains stationary (refer to Figure 4).
7. **Dormancy Transition:** If a module's own potential value becomes negative, it transitions to a dormant state. In this state, it detaches from the swarm and moves randomly within the space.
8. **Dormancy Reversal:** When a dormant module comes into contact with another active module, it reverts to an active state.

Based on the outlined algorithm, each module operates autonomously, making decisions and executing movements solely based on information gathered from neighboring modules. As evident from the process, the modules do not employ coordinate values or share a unified coordinate system. Moreover, they do not maintain the coordinates of the light source; instead, they only perceive its direction. Information exchange is restricted to neighboring modules, and the algorithm utilizes this information, along with the perceived direction of the light source and neighboring modules, to determine movement.

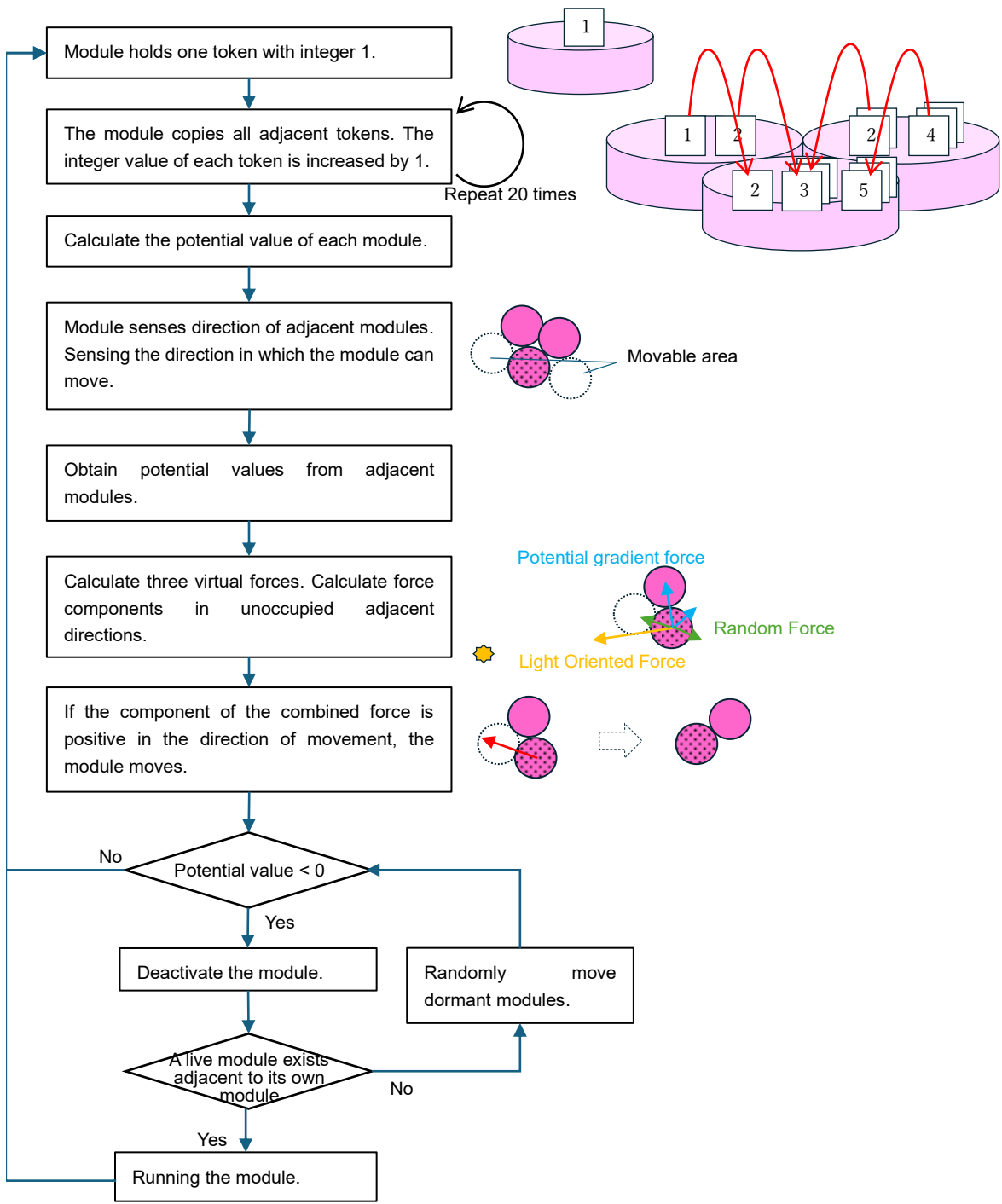

Figure 4: Flowchart of the Proposed Algorithm

## 2.4 Demonstration of Algorithm and Computational Conditions

A JavaScript program was developed to simulate the proposed algorithm. This program can be executed on commonly used web browsers through HTML files. The JavaScript program code and HTML files are provided as supplementary materials.

The program enables each module to move within the cellular space according to the algorithm described in the previous section. However, it requires parameters that determine the relative strength of the three forces acting on the module. The following parameters were set to be shared by each module:

1. **Morphological Parameter w:** This parameter, pre-recorded in each module, is utilized in Ishida's model. The value of **w** influences the transition of Turing patterns from speckled to striped patterns.
2. **Module Attraction Strength (p=0 to 16):** When **p** = 0, the force of attraction to the module is zero.
   
   **Strength of Attraction Force between Modules (Direction of Movement)** = Potential Difference from Adjacent Modules × Ratio of Directional Components × **p**
3. **Light Source Attraction Strength (l = 5, 10, 15):**
   
   **Force toward the Light Source (Direction of Movement)** = Intensity of Light in Each Cell × Ratio of Directional Components × **l**
4. **Random Movement Force (r = 1, 2, 3) = Random Force (Direction of Movement):**
   Random Number (-1, 0, 1) × **r**

The following initial conditions were established for the field:
- **Number of Active Modules:** Modules were initially placed within a circular area of radius **R** centered on a single point in space. The calculation parameters were set based on this radius **R** (a larger **R** value results in a greater number of initially placed modules).
- **Light Source Placement:** The light source could be either present or absent.
- **Obstructing Wall Placement:** Obstructing walls could be present or absent. If present, they could have either a 5-cell or 3-cell gap.
- **Pre-random Placement of Dormant Modules:** Dormant modules could be pre-randomly placed within the space with a probability of 0% to 30%.

The standard case settings were **p** = 8, **l** = 10, **r** = 1, initial radius **R** = 5.0, pre-random placement of dormant modules 0.0, and **w** = 0.2.

## 2.5 Morphological Evaluation Index

Each module is initially placed in a state of a cluster and moves while maintaining connections with its neighbors. Therefore, the cluster is essentially a single continuous group without being divided. In this study, we employed the following morphological evaluation index to assess whether this group exhibits a circular or elongated shape.

**Morphological Evaluation Index =** Sum of distances between each module and the center of gravity of the cluster / Sum of distances when the same number of modules are arranged in a circle　　(3)

The closer the cluster resembles a circular shape, the smaller the morphological evaluation index becomes. Conversely, the more spread out and elongated the cluster, the larger the morphological index becomes. A perfectly circular cluster would yield a morphological evaluation index of 1.

## 3. Results
### 3-1 Effectiveness of the Proposed Model

The proposed model, which involves mutual exchange of integers with neighboring cells before decision-making, differs from Ishida's model. It was compared the results of our model to those of Ishida's model, which is known to generate Turing patterns, to assess its effectiveness.

### (1) Potential Value Distribution in the Ishida Model

The Ishida model exhibits a characteristic distribution of potential values, which plays a crucial role in the formation of Turing patterns. The distribution of potential values in the Ishida's model exhibits a characteristic pattern: the potential value gradually increases towards the center of the Turing pattern. In the Ishida's model, the spots are located in the region where the potential value exceeds 0. At the edge of the pattern, the potential value drops to zero. Figure 5A illustrates an example of the temporal evolution of the potential distribution in the Ishida's model.

### (2) Potential Value Distribution in the Proposed Model

Figure 5B illustrates the distribution of potential values according to the proposed model. As evident in the figure, the distribution exhibits a higher center, indicating the emergence of a potential value distribution similar to that of the Ishida's model. Furthermore, it can be observed that the potential region diminishes as the value of the shape parameter **w** increases. This observation aligns with the disappearance of the pattern in the Ishida's model as the **w** value increases.

It is important to note that the proposed model's calculations are restricted to cells with modules, precluding the confirmation of pattern expansion (positive potential value) in space as observed in Ishida's model. However, by initially placing modules randomly to maximize space occupancy, a calculation approach similar to the cellular automaton model can be achieved. Under the same calculation conditions as in Figure 5B, when dormant modules are randomly distributed throughout the space with a probability of 0.3, a speckled pattern (distribution of positive potential values) akin to a Turing pattern emerges, as shown in Figure 5C.

This serves as evidence that the proposed model can implement an algorithm equivalent to the Turing pattern model. Thus, it was confirmed that the potential region can be formed solely through the exchange of numbers with neighboring cells. This implies that potential regions can be calculated without the need for module leaders, individual IDs, or coordinate information. Consequently, modules can be integrated and maintained within a single group simply by driving them to gather in the direction of their higher potentials.

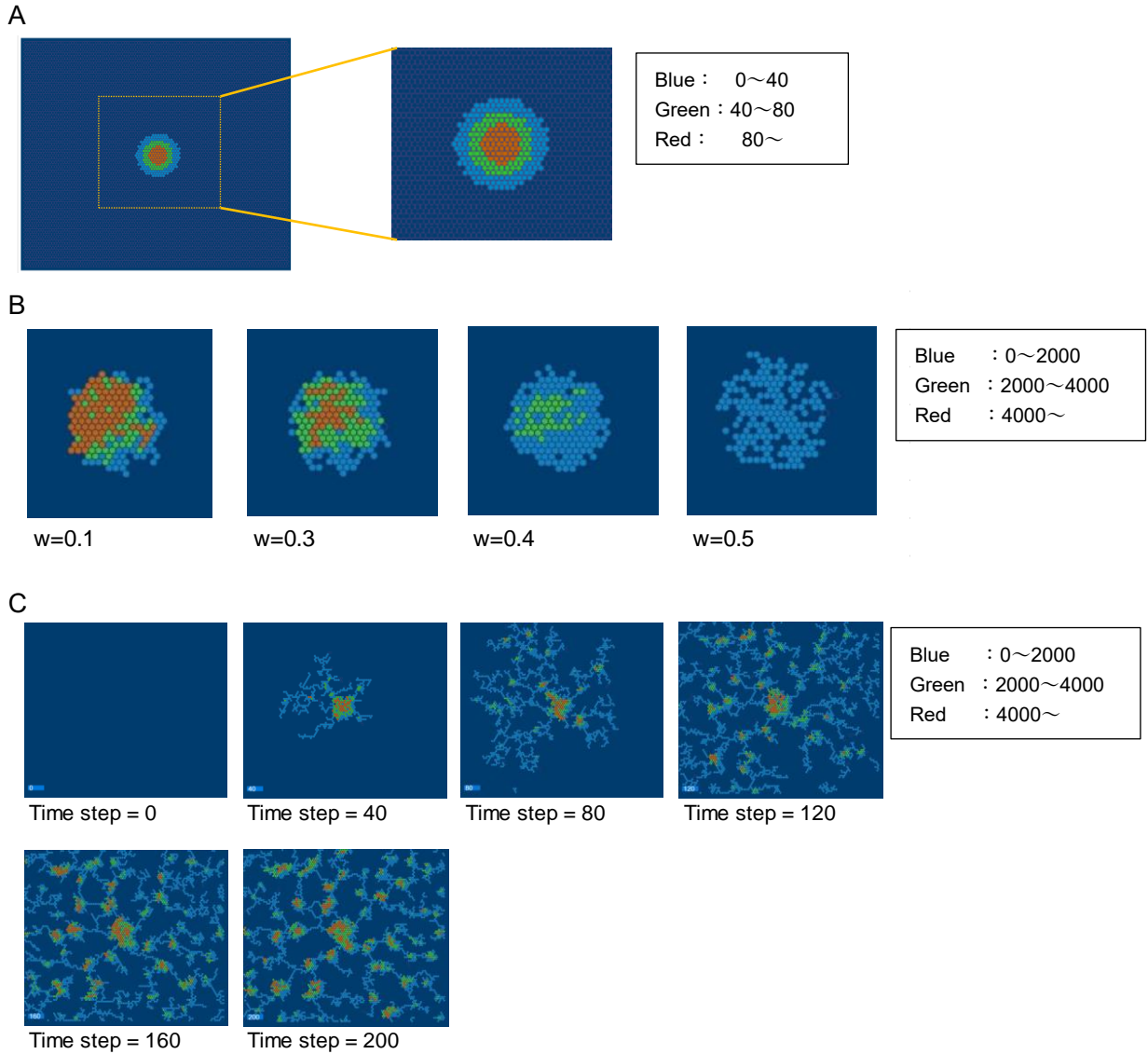

Figure 5: Examples of Potential Value Distributions **A**: Potential distribution in the Ishida longitudinal model, **B**: Calculation with the proposed model (no light source, **p** = 8, **l** = 10, **r** = 1, initial radius **R** = 7.0, **w** = 0.2, number of iterations 20), **C**: Cellular automaton-like calculation with dormant modules (white modules) filled in space (**p** = 8, **l** = 10, **r** = 1, initial radius **R** = 5.0, **w** = 0.2, dormant modules placed throughout the space with a probability of 0.3)

### 3-2 Effectiveness of the Attraction Force between Modules

The attraction force between modules plays a crucial role in the collective movement and pattern formation of the proposed model. Modules are constrained to move to adjacent vacant spaces while maintaining connections with neighboring modules, resulting in the collective movement of connected modules. The magnitude of the attraction force exerted by a module is determined by the difference in potential values between it and its neighboring modules. The combined force of these attraction forces and other forces acting on the module is calculated, and the module moves in the direction of the positive resultant force. The effect of the parameter **p**, which determines the relative strength of the attraction force between modules compared to other forces, was investigated.

Figure 6 presents the simulation results. Initially, live modules (represented by red cells) are placed at the center. The results are shown for each value of **p** at time step 200. In the case of **p** = 0, the attractive force is absent, and the modules remain connected but spread out and become disordered. Conversely, when **p** is greater than 4, the modules maintain a near-circular shape. These results demonstrate that the introduction of a non-zero force between modules (**p** > 0) is essential for maintaining group cohesion. Interestingly, the group's shape remains relatively consistent regardless of the specific value of **p**, suggesting that the magnitude of the force has a limited impact on the overall outcome.

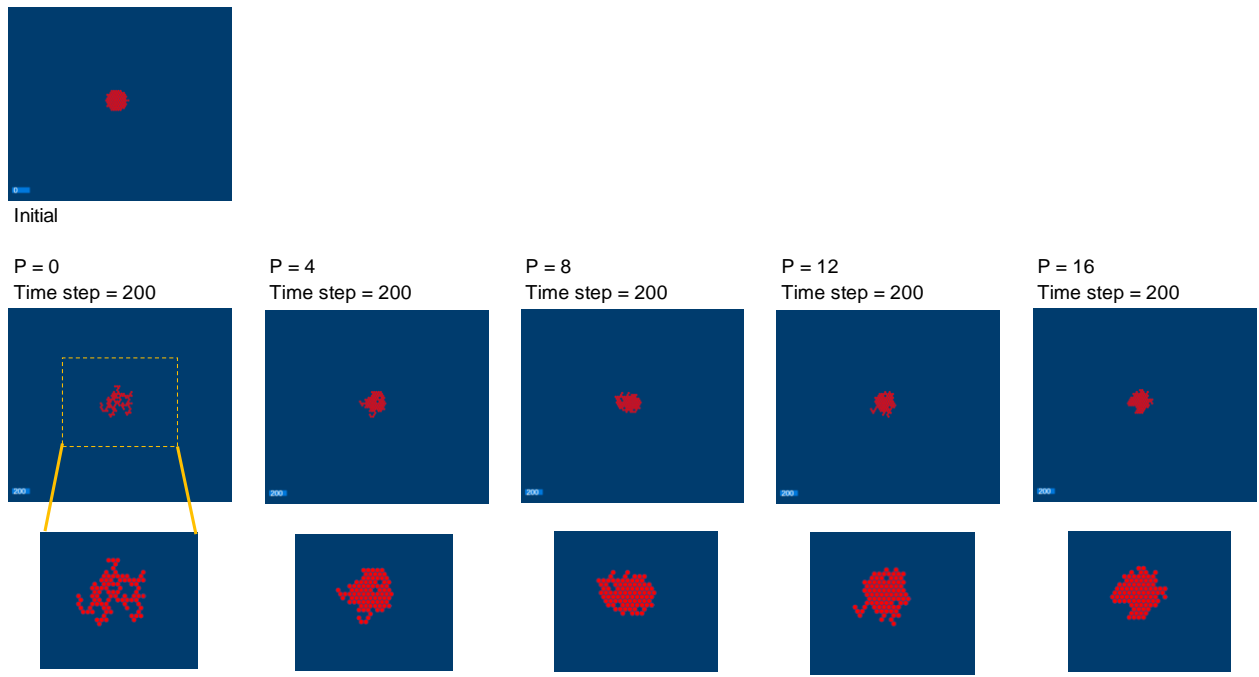

Figure 6 Effectiveness of attraction force between modules; morphology of a group of modules with parameter p (no light source, **p** = 0 to 16, **l** = 10, **r** = 3, initial radius **R** = 5.0, **w** = 0.2, number of iterations 100)

### 3-3 Light-induced movement of a group of modules

Figure 7 presents the simulation results when a light source is introduced into a space and the modules exposed to the emitted light experience an attractive force towards the light. As depicted in Figure 7A, it is assumed that light does not reach modules shadowed by other modules, and the attractive force towards light is not exerted on them. Modules lying on the straight line connecting the center of the light source cell and the center point of each module cell are considered to be in the shade. However, some modules not on the group's surface still receive light due to diffraction and reflection effects in the actual device, a reasonable assumption.

Figure 7B illustrates the time-series results of movement towards the light source for two cases: the standard case (**p** = 8) and the **p** = 0 case. In the **p** = 8 case, the modules move towards the light source while maintaining their relatively round shape. In contrast, in the **p** = 0 case, the absence of a force bringing the modules together causes the group to stretch out and move. The red cells in the figure represent cells with modules, and the total number of modules is 89.

Figure 7C shows the time series of shape evaluation values for the **p** = 0 and **p** = 8 cases. Smaller evaluation values indicate a more rounded and cohesive shape. The **p** = 8 case consistently exhibits lower shape evaluation values compared to the **p** = 0 case.

A Shadow distribution

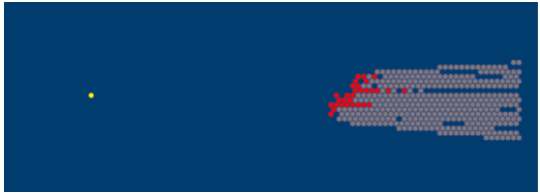

B Movement of modules group toward a light source

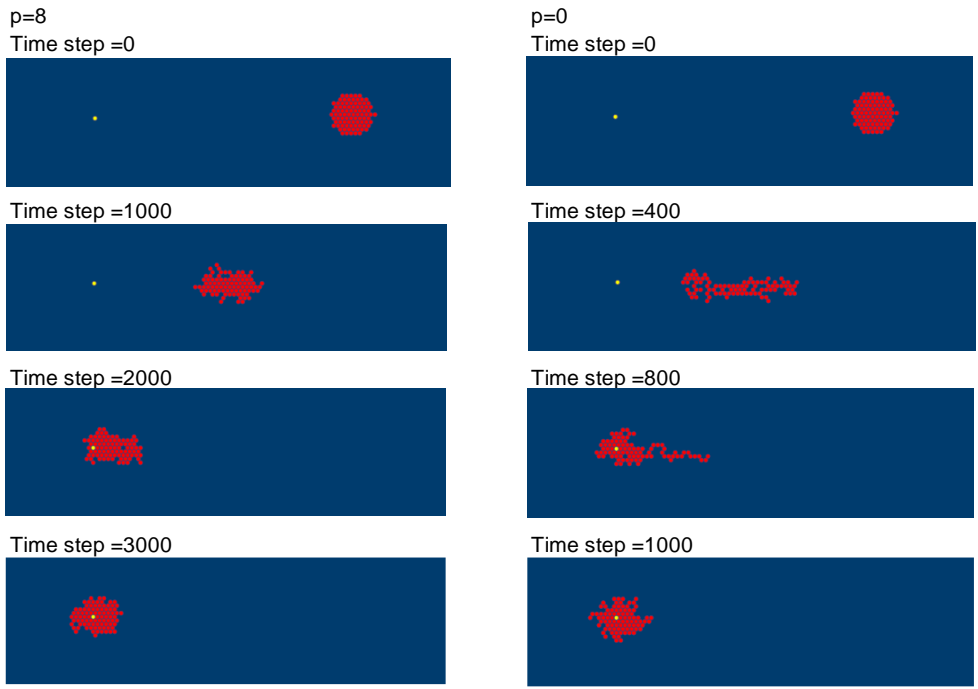

C Time series of shape evaluation values

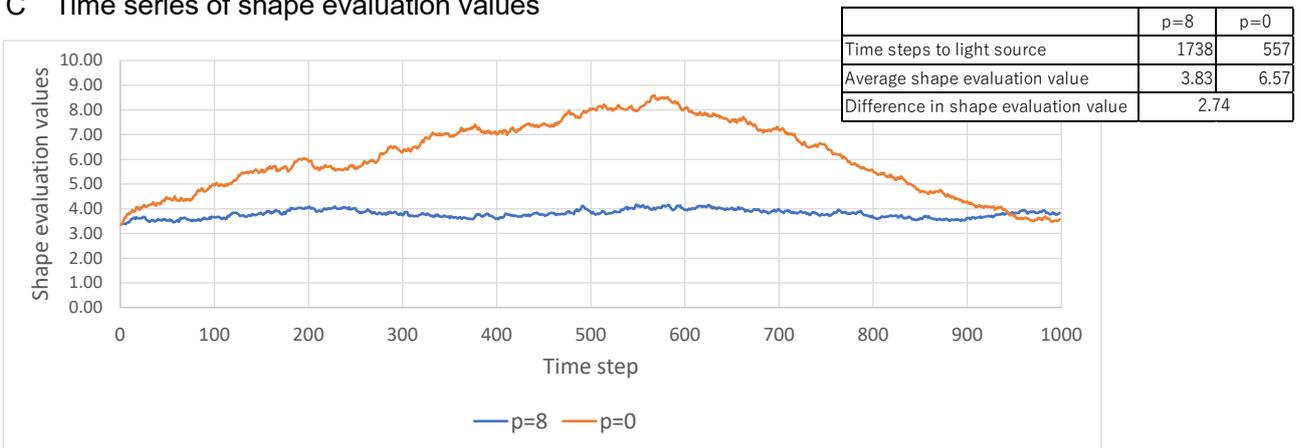

|  | p=8 | p=0 |
|---|---|---|
| Time steps to light source | 1738 | 557 |
| Average shape evaluation value | 3.83 | 6.57 |
| Difference in shape evaluation value | 2.74 | |

Figure 7 Calculated results of the movement of a group of modules to a light source. A Shadow distribution, B Calculated results of the movement of a group of modules to a light source, C Time series of shape evaluation values; smaller evaluation values indicate a more rounded and coherent shape ($\mathbf{p}$=0,8, $\mathbf{l}$=10, $\mathbf{r}$=3, initial radius $\mathbf{R}$=5.0 (number of modules is 89), $\mathbf{w}$=0.2).

Table 1 summarizes the simulation results for varying initial module placement radii (**R**). When **R** is between 2 and 5, the modules maintain their circular shape as **R** increases, with larger values of **R** resulting in slower movement towards the light source. For **R** values of 6 or greater, the modules lose their circular shape and elongate, with the tip reaching the light source more quickly. The average shape evaluation index remains relatively constant across different numbers of modules.

Table 1 Results with the initial modules placement radius **R**;

|  | R=2 | R=3 | R=4 | R=5 | R=6 | R=7 | R=8 |
|---|---|---|---|---|---|---|---|
| Number of modules | 17 | 33 | 61 | 89 | 129 | 179 | 235 |
| Number of steps to reach light source | 1484 | 1544 | 1393 | 1804 | 1682 | 1680 | 1572 |
| Average shape evaluation value | 1.28 | 1.15 | 1.28 | 1.14 | 1.20 | 1.24 | 1.20 |

**Table 2: Simulation results for varying random force parameter (r)**

|  | r=0 | r=3 | r=5 |
|---|---|---|---|
| Number of modules | 89 | 89 | 89 |
| Number of steps to reach light source | 1819 | 1913 | 1804 |
| Average shape evaluation value | 1.16 | 1.16 | 1.14 |

Table 2 presents the simulation results for different values of the random force parameter (**r**), which determines the relative magnitude of the random force acting on the modules. The parameter **r** sets the strength of the random force acting on the modules along the direction of the vacant destination axis during their movement. As can be observed from Table 2, altering the parameter r has a minimal impact on both the arrival step and the shape evaluation index.

Table 3 summarizes the simulation results for varying values of the light source force parameter (**l**), which determines the relative strength of the force pulling the modules towards the light source. As the value of **l** increases, the force towards the light source becomes stronger, resulting in the modules reaching the light source more quickly. However, a larger value of **l** also tends to stretch the group of modules and increase the shape evaluation parameter.

**Table 3: Simulation results for varying light source force parameter (l)**

|                                    | l=5  | l=10 | l=15 |
|------------------------------------|------|------|------|
| Number of modules                  | 89   | 89   | 89   |
| Number of steps to reach light source | 4019 | 2021 | 1160 |
| Average shape evaluation value     | 1.08 | 1.15 | 1.34 |

Figure 8 illustrates the simulation results for the movement of a group of modules towards the light source in the presence of an obstacle (wall). The scenarios depicted involve wall gaps of 3 and 5 cells, respectively. In both cases, the modules successfully navigate through the gap and reach the light source. Despite experiencing significant deformation while traversing the gap, the modules ultimately achieve their objective.

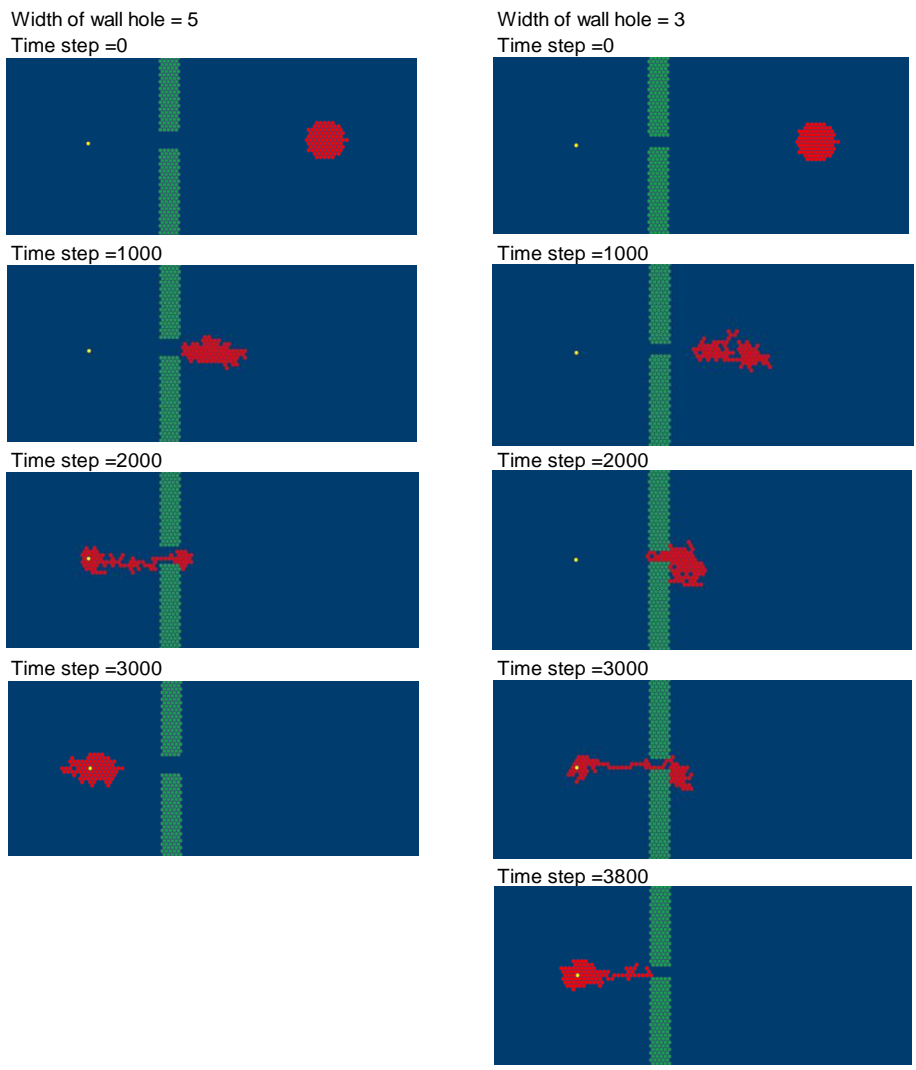

Figure 8 Simulation results for the movement of a group of modules towards a light source in the presence of an obstacle (wall) under standard conditions ($p = 8$, $l = 10$, $r = 3$, initial radius $R = 5.0$, $w$

= 0.2). The scenarios depicted involve wall gaps of 5 and 3 cells, respectively. When the wall gap size was 5 cells, the modules required 1630 steps to reach the light source. As the gap narrowed to 3 cells, the number of steps increased to 2522.

### 3.4 Growth and Replication Patterns of Module Groups

Figure 9 illustrates the simulation results for the scenario where no light source is present. Initially, dormant modules are randomly distributed throughout the space with a placement probability of 0.1 under standard conditions. Dormant modules undergo random movement and are converted into live modules upon contact with live modules, enabling the observation of a gradual module growth pattern.

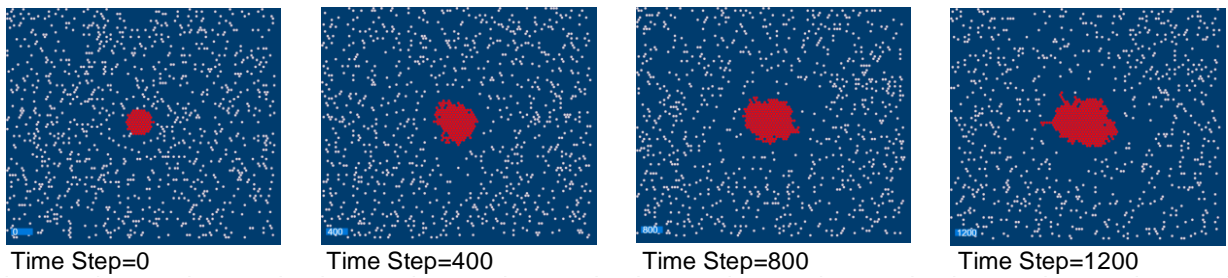

| Time Step=0 | Time Step=400 | Time Step=800 | Time Step=1200 |

Figure 9 Module growth pattern; no light source, standard parameters set (p = 8, l = 10, r = 1, initial radius R = 5.0, w = 0.2), set dormant modules throughout the space with 0.1 probability.

By introducing a probability for modules to transition into a dormant state even when the potential value exceeds zero, we can simulate the effect of module disconnection. Figure 10A presents the simulation results for the growth pattern of dormant modules placed with a probability of 0.1. As can be observed, the modules exhibit a splitting behavior. This phenomenon arises from the mesh-like expansion of the potential value distribution, as illustrated in Figure 5C, which drives the group of modules to expand into regions with positive potential values, forming a speckled pattern. Increasing the initial placement probability of dormant modules to 0.2 and 0.3 leads to the formation of a spotted pattern with fewer time steps. These results demonstrate that the initial placement of dormant modules as a "resource" leads to the formation of Turing-like patterns. Furthermore, by adjusting the value of the shape parameter **w**, a reticulate morphology can be generated, as shown in Figure 10B. Such observations suggest that the creation of modules with similar functionalities at the micro-nano scale could potentially lead to module replication and the emergence of network structures, opening up avenues for further exploration.

A Self-replication pattern of modules (time series results)

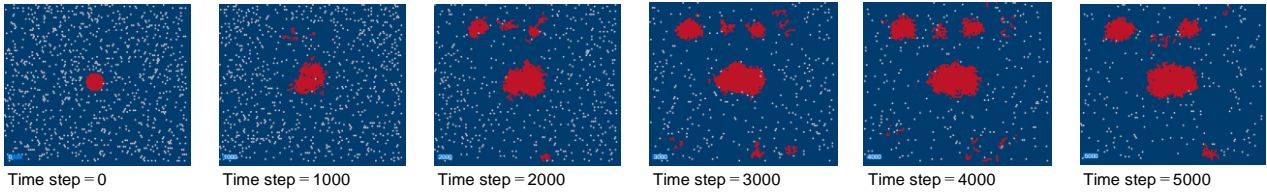

B Self-replication pattern of modules with placement rate of dormant modules

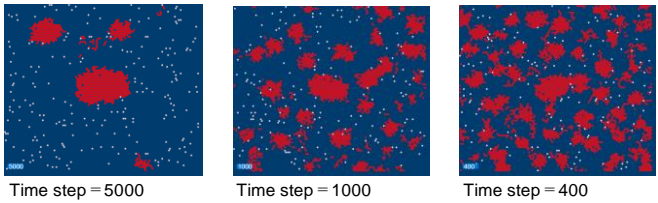

C Result of emergent patterns with shape parameter w

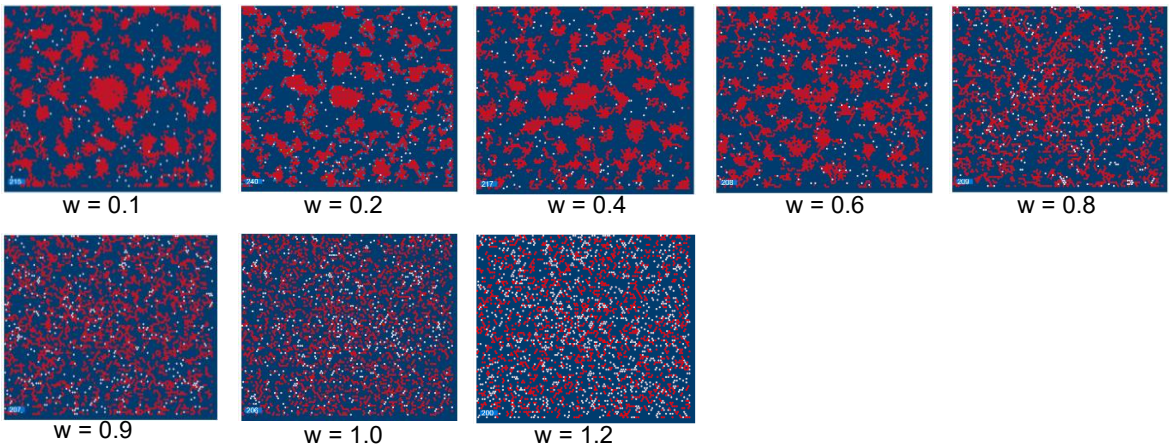

Figure 10 Self-replicating pattern of module groups; A Time series results ($p = 8, l = 10, r = 1$, initial radius $R = 3.0$, $w = 0.2$, module groups separated with probability 1/7, dormant modules placed throughout space with probability 0.1), B Results with shape parameter $w$ (dormant modules placed throughout space with probability 0.3).

## 4. Discussion

In contrast to many conventional swarming robots that require each module to possess the coordinate data of the desired shape and a designated seed module, the proposed model utilizes an algorithm based on Turing patterns to achieve the integration and movement of modules without employing a coordinate system. This approach relies solely on the exchange of information between neighboring modules. The proposed model offers several advantages:

(1) **Elimination of Design Data and Coordinate Information:** The formation of a circular flock does not necessitate any specific design data or coordinate information.

(2) **Absence of Coordinate System Maintenance:** Maintaining a coordinate system is unnecessary. Instead, the modules only need to perceive the direction of light and the direction of adjacent modules.

(3) **Localized Information Exchange:** Information acquisition from distant modules is not required. The exchange of information with neighboring modules is sufficient.

(4) **Adaptability to Light Source Movement and Gap Traversal:** The modules can navigate towards the light source or pass through gaps while maintaining the integrity of the swarm.

The proposed model enables each module to maintain a cohesive group and act based on very simple information exchange and decision-making processes. This simplicity stands in contrast to the void model [44], which requires each module implementing the void model to measure and determine the position, direction, and distance of each neighboring module independently. This demand for high perceptual ability adds complexity to the void model. In contrast, the proposed algorithm offers a simpler approach that could contribute to the advancement of swarm robot control.

The proposed algorithm is particularly relevant for building micro- and nano-level swarm robots, where the small scale of the modules limits the functionalities that can be incorporated into each module. This algorithm, which can achieve swarm movement with a simple procedure, has the potential to enhance the engineering feasibility of swarm robots at the micro- and nano-level.

One potential drawback of the algorithm is that if too many modules are gathered together, the group's movement may slow down or even stall altogether. To address this issue, it would be necessary to investigate the optimal number of modules based on the specific task assigned to the group.

# 5. Conclusion

Despite its simplicity, the proposed algorithm has demonstrated the ability to generate swarm behavior reminiscent of living organisms. It has been shown that a group of robots can be effectively controlled through a basic process where each module calculates potential values and adjusts its movement based on those values.

While not all parameter combinations have been investigated, future work should focus on identifying other potentially useful parameter combinations. Additionally, it would be worthwhile to explore parameters that could elongate the modules in a specific direction to create a long robot shape.

Currently, the algorithm is limited to creating swarms of modules of a single type. To address this, future research will focus on developing a model that allows for the emergence of hierarchical structures within the swarm. This involves incorporating the other Ishida model [44], which combines Turing patterns and the game of life, to create even more diverse patterns and potentially enable the emergence of hierarchical, limbed structures. Details of this approach will be presented in a separate paper.

While the proposed algorithm has been shown to be effective in a discrete space, further validation is required in a continuous space. Additionally, building a prototype to demonstrate the model on an actual machine would be a valuable step towards real-world implementation.

## Supplementary files

p 力が0の場合.mp4

Video 1 Effectiveness of attraction force between modules; morphology of a group of modules with parameter p (no light source, $\underline{p = 0}$, l = 10, r = 3, initial radius R = 5.0, w = 0.2)

p 力が8の場合.mp4

Video 2 Effectiveness of attraction force between modules; morphology of a group of modules with parameter p (no light source, $\underline{p = 8}$, l = 10, r = 3, initial radius R = 5.0, w = 0.2)

光源への移動（標準ケース）.mp4

Video 3 Calculation results of the movement of a modules group toward a light source(p = 8, l = 10, r = 3, initial radius R = 5.0, w = 0.2)

光源への移動（P＝0ケース）.mp4

Video 4 Calculation results of the movement of a modules group toward a light source(p = 0, l = 10, r = 3, initial radius R = 5.0, w = 0.2)

壁有の動画（すきま5）.mp4

Video 5 Calculated results of the movement of a group of modules to a light source when a wall is present; the results are for the standard case (p = 8, l = 10, r = 3, initial radius R = 5.0, w = 0.2, range = 0) with a wall.

複製パターン初期パーティクル2.mp4

Video 6 Self-replicating pattern of module groups (p = 8, l = 10, r = 1, initial radius R = 3.0, w = 0.2, module groups separated with probability 1/7, dormant modules placed throughout space with probability 0.1)

## Statements & Declarations

Author Contributions: The author, TI, conducted all the research.

Funding: This research was supported by grants from Japan Society for the Promotion of Science, KAKENHI Grant Number 19K04896.

Data Availability Statement: The source code for the computational model of this study is available on GitHub at https://github.com/Takeshi-Ishida/Algorithm-applied-the-Turing-pattern-model-to-control-active-swarm-robots

Acknowledgments: A preprint has previously been published on the following arXiv site: https://・・・・.

Conflicts of Interest: The author declares no conflicts of interest regarding the publication of this paper.